\documentclass[runningheads]{llncs}
\usepackage{graphicx}
\usepackage{gensymb}
\usepackage{subfig}
\usepackage{multirow}
\usepackage{hyperref,xcolor}

\begin{document}
\title{Towards Real-Time Head Pose Estimation: Exploring Parameter-Reduced Residual Networks on In-the-wild Datasets}
\author{Ines Rieger \and
	Thomas Hauenstein \and
	Sebastian Hettenkofer \and
	Jens-Uwe Garbas}
\authorrunning{I. Rieger et al.}
\titlerunning{Towards Real-Time Head Pose Estimation}
%
\institute{Fraunhofer-Institute for Integrated Circuits IIS, \\ Am Wolfsmantel 33, 91058 Erlangen, Germany \\ \email{ines.rieger@iis.fraunhofer.de} \\}
\maketitle              
\begin{abstract}
Head poses are a key component of human bodily communication and thus a decisive element of human-computer interaction. Real-time head pose estimation is crucial in the context of human-robot interaction or driver assistance systems. The most promising approaches for head pose estimation are based on Convolutional Neural Networks (CNNs). However, CNN models are often too complex to achieve real-time performance. To face this challenge, we explore a popular subgroup of CNNs, the Residual Networks (ResNets) and modify them in order to reduce their number of parameters. The ResNets are modified for different image sizes including low-resolution images and combined with a varying number of layers. They are trained on in-the-wild datasets to ensure real-world applicability. As a result, we demonstrate that the performance of the ResNets can be maintained while reducing the number of parameters. The modified ResNets achieve state-of-the-art accuracy and provide fast inference for real-time applicability.
\keywords{Head pose estimation \and Residual Network \and Real-time.}
\end{abstract}
\section{Introduction}
Head poses are a key aspect of human non-verbal communication. As a consequence, automatic head pose estimation plays an important role in human-computer interaction. Several use cases in real-world scenarios include head pose estimation: In autonomous driving, head poses are used to estimate the driver's level of attention. Inattentive drivers can then be encouraged to focus on the road again \cite{kuwahara2018driving}. In order to diminish the risk of collisions, the attention level of the surrounding pedestrians can also be estimated from their head poses \cite{benenson2014ten,geronimo2010survey}. Head pose estimation is further used as one of the key aspects in real-time human-robot interaction, e.g. in domestic environments \cite{van2011head} to provide a natural interaction mode with its users. In behavioural studies, head poses can be used to identify social groups \cite{leach2014detecting} or a person's target of interest \cite{leroy2013second}. Head poses are also part of the Facial Action Coding System (FACS) \cite{friesen1978facial} to decode emotions and therefore contribute to the interpretation of facial expressions \cite{izard2013human}. Thus, real-time head pose estimation is crucial for several real-world applications, but still faces challenges such as slow inference time or robustness for these settings. This paper aims at providing a real-time solution by exploring small Convolutional Neural Network (CNN) models for low resolution input images trained on in-the-wild datasets.\\
\indent CNNs, a specialized kind of feed-forward neural network, have proven to be advantageous for various image and video processing tasks such as object detection or object recognition \cite{lecun2015deep}. One particular successful CNN architecture is the Residual Network (ResNet) architecture \cite{he2016deep}, which provides effective training with very deep networks through shortcut connections. The shortcut connections enclose blocks of stacks of convolutional layers and enable a second way to propagate information forward and backward through the network. Veit et al. \cite{veit2016residual} proved that not all blocks contribute equally to the learning process  by investigating the gradient flow. Presumably, a more shallow ResNet architecture could learn the same representations and perform as well as a deeper ResNet architecture. Based on this assumption and the overall success of ResNets, this paper contributes to the field of real-time head pose estimation by exploring various modified ResNets. We can summarize our key contributions as follows:
\begin{itemize}
	\item We start to reduce the model parameters by adapting the original ResNet architecture for training with images of $112$ x $112$ pixels instead of $224$ x $224$ pixels. We further reduce the parameters by adapting the 18-layer ResNet for low-resolution images of $64$ x $64$ pixels. This 18-layer ResNet contains less parameters than the 18-layer ResNet originally proposed by He et al. \cite{he2016deep}.
	\item The modified ResNets are evaluated on two in-the-wild datasets: The \textit{Annotated Facial Landmarks in the Wild} (AFLW) dataset \cite{koestinger2011annotated} and the \textit{Annotated Faces in the Wild} (AFW) benchmark dataset \cite{zhu2012face}. In-the-wild datasets ensure real-world applicability, which is important for use cases such as driver assistance systems or human-robot interaction. These datasets include no depth information. 
	\item The performance of the implemented ResNets is evaluated with a five-fold cross-validation on the AFLW dataset and with a five time training-testing cycle on the AFW dataset. Multiple training cycles not only contribute to the robustness of the results, but also mitigate the non-deterministic behaviour of multi-thread training on GPUs. The results are measured in mean absolute error and accuracy.
	\item We compute the number of parameters and measure the inference time on a CPU and on a GPU. Low model complexity and the corresponding fast inference time is important for real-time applications.
\end{itemize}
\indent \indent The ResNet models are trained to estimate the head poses represented by Euler Angles, which measure the orientation of a rigid body in a fixed coordinate system \cite{diebel2006representing}.
\section{Related Work}
Head pose estimation approaches can be grouped in appearance-based methods, model-based methods and nonlinear regression methods.\\
\indent Appearance-based methods compare new head images with a set of exemplary, annotated heads  and pick the most similar one \cite{beymer1994face,niyogi1996example}. Despite the advantage of a simple implementation and an easy extension for new heads, there is a huge disadvantage: The method is based on the premise that similar images also have similar head poses, and thus ignore the impact of identity.\\
\indent In contrast to appearance-based methods, model-based methods follow a geometric approach by not taking the whole face into account, but only certain facial key-points. One approach uses the POSIT algorithm \cite{dementhon1995model} to fit an averaged 3-dimensional facial model onto a 2-dimensional face image annotated with facial key-points, and then computes the head pose \cite{koestinger2011annotated}. Another approach is to measure the distance of the facial key-points of the 2-dimensional image to a reference coordinate system \cite{li2014central}. The drawback of the model-based approaches is the need of high accuracy in the facial key-point detection. Estimating head poses from images with occluded face regions is therefore difficult.\\
\indent To cover the complex feature space required for head pose estimation in images in in-the-wild settings, nonlinear regression methods can present a solution. The first nonlinear regression methods for head pose estimation were support vector regression \cite{li2000support}, random forests \cite{fanelli2011real,fanelli2013random} and  multilayer perceptrons (MLP) \cite{schiele1995gaze,stiefelhagen2004estimating}. With the rise of computational power, CNNs emerged around 2007 in the field of image based head pose estimation. In contrast to MLPs, CNNs display a high tolerance to shift and distortion variance. There are several recent approaches that employ the in-the-wild datasets AFLW and AFW for training and use the AFW dataset as a testing benchmark: Patacchiola and Cangelosi \cite{patacchiola2017head} compare various LeNet-5 \cite{lecun1998gradient} variants trained with different gradient and adaptive gradient methods. Ruiz et al. \cite{ruiz2018fine} train a 50-layer ResNet with a combined loss function of mean squared loss and cross entropy loss for all three angles. They achieve good results and outperform Patacchiola and Cangelosi. Kumar et al. \cite{kumar2017kepler} use a Heatmap-CNN (H-CNN) that learns local and global structural dependencies for detecting facial landmarks and estimating the head pose. The H-CNN includes Inception modules \cite{szegedy2015going} that consist of parallel threads of stacked convolutional layers and therefore display an architecture similar to ResNets. Hsu et al. \cite{hsu2018quatnet} train their multi-loss CNN based on a combined L2 loss regression and ordinal regression loss. To counteract the \textit{gimbal lock} \cite{lepetit2005monocular}, an ambiguity problem in the Euler angle representation, they use quaternions as head pose representation. Hsu et al. \cite{hsu2018quatnet} find that their pretrained Quaternion Net outperforms their network using Euler Angles for head pose representation. Wu et al. \cite{wu2018simultaneous} train their combined face detection network on an augmented AFLW dataset combined with an unreleased own head pose dataset. An evaluation on the AFW dataset achieves state-of-the-art performance. Zhang et al. \cite{zhang2018cross} use a cross-cascading regression network with two submodules, one for facial landmark detection and one for head pose estimation and achieve state-of-the-art performance on the AFLW dataset.
\section{Residual Networks (ResNets)}
The ResNet is one of the most popular architectures for image processing with very deep neural networks. The architecture was proposed by He et al. \cite{he2016deep}, who won benchmark competitions like the ImageNet Large Scale Visual Recognition Competition (ILSVRC) 2015\footnote{\url{http://image-net.org/challenges/LSVRC/2015/}, accessed 14.12.2018.} and the Common Objects in Context (COCO)\footnote{\url{http://cocodataset.org/\#detections-challenge2015}, accessed 14.12.2018.} competition with an ensemble of ResNets. They have also proven the successful training with over one thousand layers \cite{he2016deep}.

\subsection{Original ResNets}
\indent ResNets use the concept of shortcut connections with the effect that the input of the subsequent layers does not only contain the information of the immediate preceding layer, but from all preceding layers. This is contradictory to hierarchical CNNs, where the input of the layers does only contain information of the directly preceding layer. The shortcut connections can resolve the problem of vanishing or exploding gradients and the degradation of the training error in hierarchical CNNs. The degradation of the training error describes that the training and test error increases, when the network depth grows.\\
\indent A shortcut connection is called an \textit{identity shortcut}, when the input and output dimension stays the same within a block. Identity shortcuts do not introduce additional parameters in the network. When the dimension increases, He et al. \cite{he2016deep} consider two options: (1) Identity mapping with extra zero entries or (2) projection shortcuts using $1$ x $1$ convolutions. As ResNets do not use as many filters as classic CNNs to achieve the same depth, they are parameter-reduced.\\
\indent The residual block (see Fig.~\ref{fig:ResNet} (right side)) described by He et al. \cite{he2016deep} uses the Rectified Linear Unit (ReLU) activation function with a preceding batch normalization layer. The shortcut connection encloses a block of two convolutional layers $F(x_l)$, where the input $x_l$ is added to the result of the two convolutional layers, resulting in $x_{l+1}$ in the forward propagation (see Eq.~\ref{eq:identitymappingNetwork_3}). This equation does not hold for blocks using the projection shortcut, but since the likelihood for such blocks is low, He et al. \cite{he2016deep} do not expect this to have a great impact.
\begin{equation}
x_{l+1} = x_l + F(x_l)
\label{eq:identitymappingNetwork_3}
\end{equation}
\indent 
In the backpropagation, the incoming gradient is split into two additive terms: The error is propagated through the shortcut connection and through the residual function, where the weights of the convolutional layers are adjusted. Since any gradient is a summation, they are not likely to vanish.\\
\indent To enhance the performance of the ResNet, He et al. \cite{he2016identity} propose a pre-activated residual block, where the Batch Normalization layer is placed before instead of after the convolutional layer. The presented experiments in this paper use the pre-activated form of the residual block.
\subsection{Modified ResNets}
The challenging aspect of real-time applications is the complexity of the trained models and the resulting computing time. To address this challenge, we explore three parameter-reduced ResNets  of different depths and for different image sizes.\\
\indent These ResNets (see Table \ref{Tab:ResNets}) are modified to process images with a lower resolution of $112$ x $112$ pixels and $64$ x $64$ pixels instead of $224$ x $224$ pixels as in the original ResNet \cite{he2016identity}. As a first step, the stride of the first convolutional layer is changed to one, so the size of the feature map is not reduced. This is different to the original ResNet, where a stride of two is used. Consequently, as the convolutional layer does not reduce the size of the feature map, the ResNet can process input images with a size of $112$ x $112$ pixels instead of $224$ x $224$ pixels. The ResNet34-112 and ResNet18-112 are modified to take $112$ x $112$ pixels as input. Their layers are divided in four stacks similar to \cite{he2016deep}. The smallest proposed ResNet of He et al. \cite{he2016deep} is a ResNet with 18 layers divided in four stacks. To further reduce the parameters, we propose the ResNet18-64, which uses only three stacks instead of four stacks. This allows low resolution inputs of $64$ x $64$ pixels while significantly decreasing the number of parameters.
\renewcommand{\arraystretch}{1.1}
\begin{table}[h!]
	\caption{Overview of modified ResNets}
	\begin{tabular}{|p{3 cm}|p{3.0cm}| p{1.5cm}|p{1.5cm}|p{2cm}| }
		\hline
		\textbf{ResNet Model}& \textbf{Input Size}  & \textbf{Stacks} & \textbf{Layers} &\textbf{Parameters}  \\
		\hline
		\textbf{ResNet34-112} &  $112$ x $112$ pixels & [3,4,6,3] &  34  & 21.27 x $10^6$\\
		\textbf{ResNet18-112}&  $112$ x $112$ pixels&  [2,2,2,2] & 18 & 11.17 x $10^6$ \\
		\textbf{ResNet18-64}&  $64$ x $64$ pixels & [2,3,3,0] & 18 & 4.25 x $10^6$\\
		\hline
	\end{tabular}
	\label{Tab:ResNets}
\end{table}
\begin{figure}[h!]
	\includegraphics[width=\textwidth]{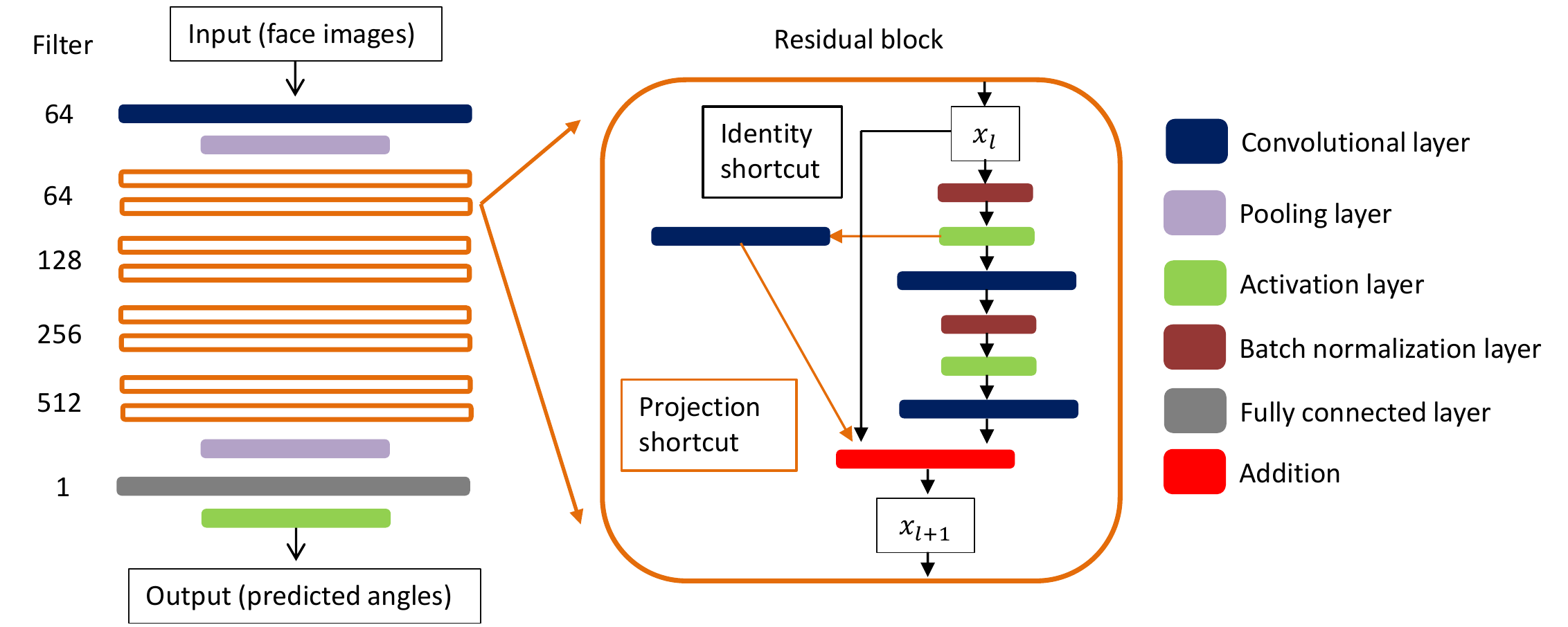}
	\caption{Modified ResNet architecture with 18 layers divided in four stacks for inputs of $112$ x $112$ pixels (ResNet18-112).} 
	\label{fig:ResNet}
\end{figure}\\
\indent The modified ResNets (see Fig.~\ref{fig:ResNet}) use pre-activated residual blocks with projection shortcuts for an increase in dimensionality and identity shortcuts, when the dimension stays the same. In contrast to the original ResNet \cite{he2016identity}, the hyperbolic tangent function is used as the activation function instead of the ReLU function. The tangent function computes values in the range of [-1,1] and the labels and image pixel values are normalized to values in this range as well. Furthermore, the projection shortcuts for increasing the dimensions are placed after instead of before the first set of batch normalization and tangent activation layer, as the batch normalization provides a regularization effect of the image pixel values.
\section{Datasets}
To train the models for real-life settings, two in-the-wild datasets are used for training and testing.
\subsection{Annotated Facial Landmarks in the Wild (AFLW)}\label{sec:aflw}
The AFLW dataset \cite{koestinger2011annotated} provides a large variety of different faces with regard to ethnicity, pose, expression, age, gender and occlusion. The faces are in front of natural background under varying lighting conditions. The dataset contains 25,993 annotated faces in 21,997 images. The license agreement does not allow publication of the AFLW database.\footnote{\url{https://www.tugraz.at/institute/icg/research/team-bischof/lrs/downloads/aflw/}, accessed 26.03.2019.} All images are annotated with face coordinates and the three angles yaw, pitch and roll. 56\% of the faces are tagged as female and 44\% are tagged as male. Koestinger et al. \cite{koestinger2011annotated} state that the rate of non-frontal faces of 66\% is higher than in any other dataset. The distribution of poses of the AFLW dataset is not uniform, showing fewer images with a strong head rotation. The yaw angle has a range from $-125.1\degree$ to $168.0\degree$, the pitch angle from $-90.0\degree$ to $90.0\degree$ and the roll angle from $-178.2\degree$ to $179.0\degree$. The head poses were computed with the POSIT algorithm using manually annotated facial key-points \cite{dementhon1995model}. However, it is worth noting that the resulting head poses were not manually verified. The AFLW dataset has extremely wide ranges for all angles, which supersede realistic head movements by far \cite{ferrario2002active}. 
\subsection{Annotated Faces in the Wild (AFW)}
The AFW dataset \cite{zhu2012face} shows a wide variety of ethnicity, pose, expression, age, gender and occlusion. The license agreement does not allow publication of the AFW database. The faces are positioned in front of natural cluttered backgrounds. There are 468 faces in 205 images. An annotation of the angles yaw, pitch and roll as well as face coordinates are provided. The yaw angle has a range from $-105\degree$ to $90\degree$, the pitch angle from $-45\degree$ to $30\degree$ and the roll angle from $-15\degree$ to $15\degree$, all annotated manually in steps of $15\degree$. Since the yaw angle has the widest range, this angle is normally used when testing with this dataset.
\section{Experiments}
In this section, we describe the pre-processing, training parameters, evaluation methods and results of the trained models including a comparison to other state-of-the-art approaches regarding the performance and number of parameters.
\subsection{Pre-processing}
As explained in Section \ref{sec:aflw}, some samples in the AFLW dataset are annotated with unrealistic values. Furthermore, few images are provided for extreme angles. Following the approach of Patacchiola and Cangelosi \cite{patacchiola2017head}, we filter the dataset and only keep images in the following label ranges:  $\pm100\degree$ for the yaw angle, $\pm45\degree$ for the pitch angle and $\pm25\degree$ for the roll angle. The yaw angle of the AFW dataset is also restricted to $\pm100\degree$, as this angle is used for testing the trained networks. Both datasets are converted to greyscale. We crop the images using the annotated face coordinates. Each image is scaled down to $112$ x $112$ pixels and $64$ x $64$ pixels respectively. Face images smaller than the required size are left out. For the AFW dataset, the use of face images greater than 150 pixels is an additional constraint, following the protocol in \cite{zhu2012face}. To normalize the values, the labels are rescaled from $[-100,100]$ to $[-1,1]$ and the pixel values are rescaled from $[0,255]$ to $[-1,1]$.\\ 
\indent In total, four datasets are prepared for our training: AFLW-112, AFLW-64, AFW-112 and AFW-64. The total amount of face images is 16,931 in AFLW-112, 20,872 in AFLW-64, 325 in AFW-112 and 352 in AFW-64.
\subsection{Methods} \label{methods}
\begin{figure} [h!]%
	\subfloat[ResNet18-112]{{\includegraphics[width=5.5cm]{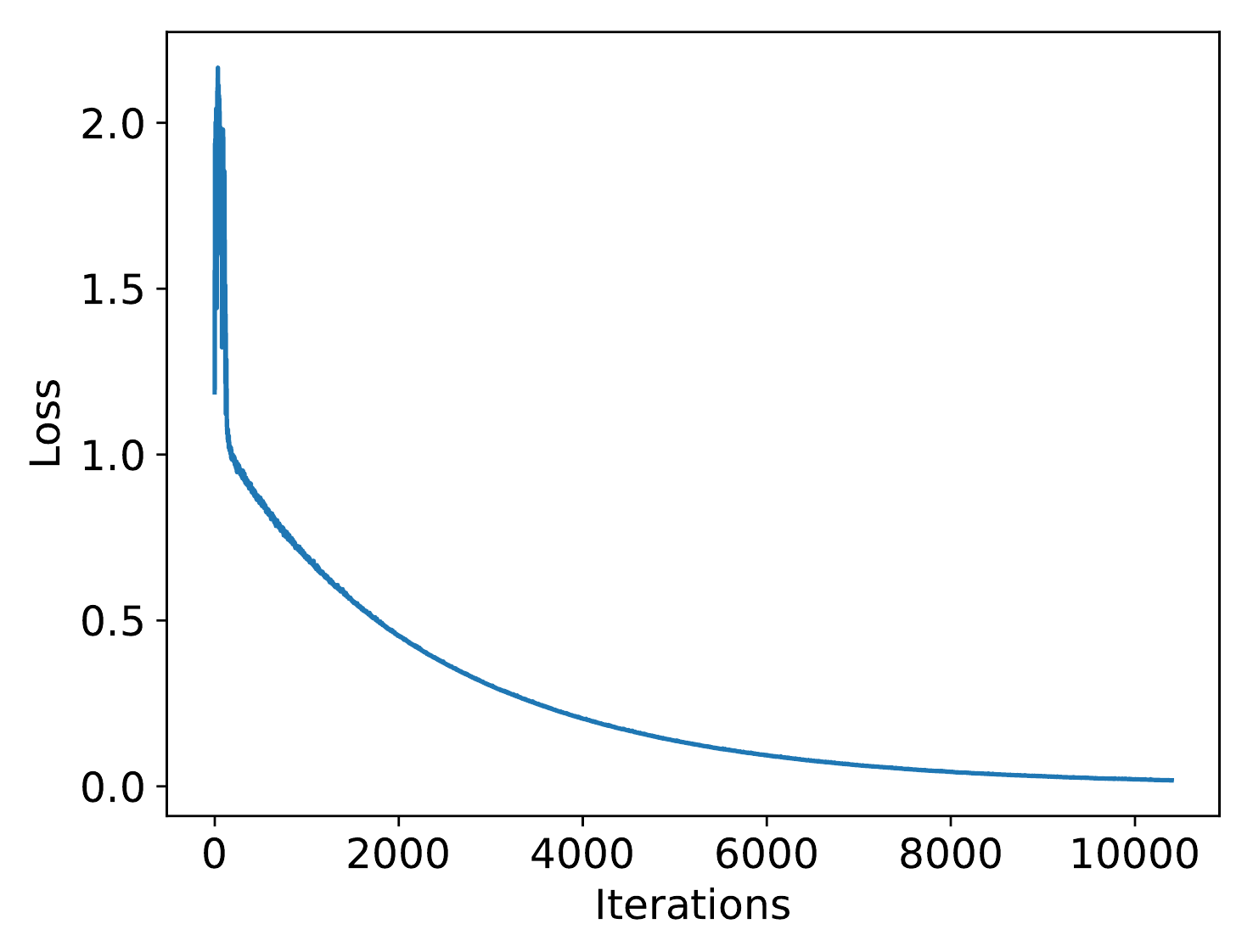} }}%
	\qquad
	\subfloat[ResNet18-64]{{\includegraphics[width=5.5cm]{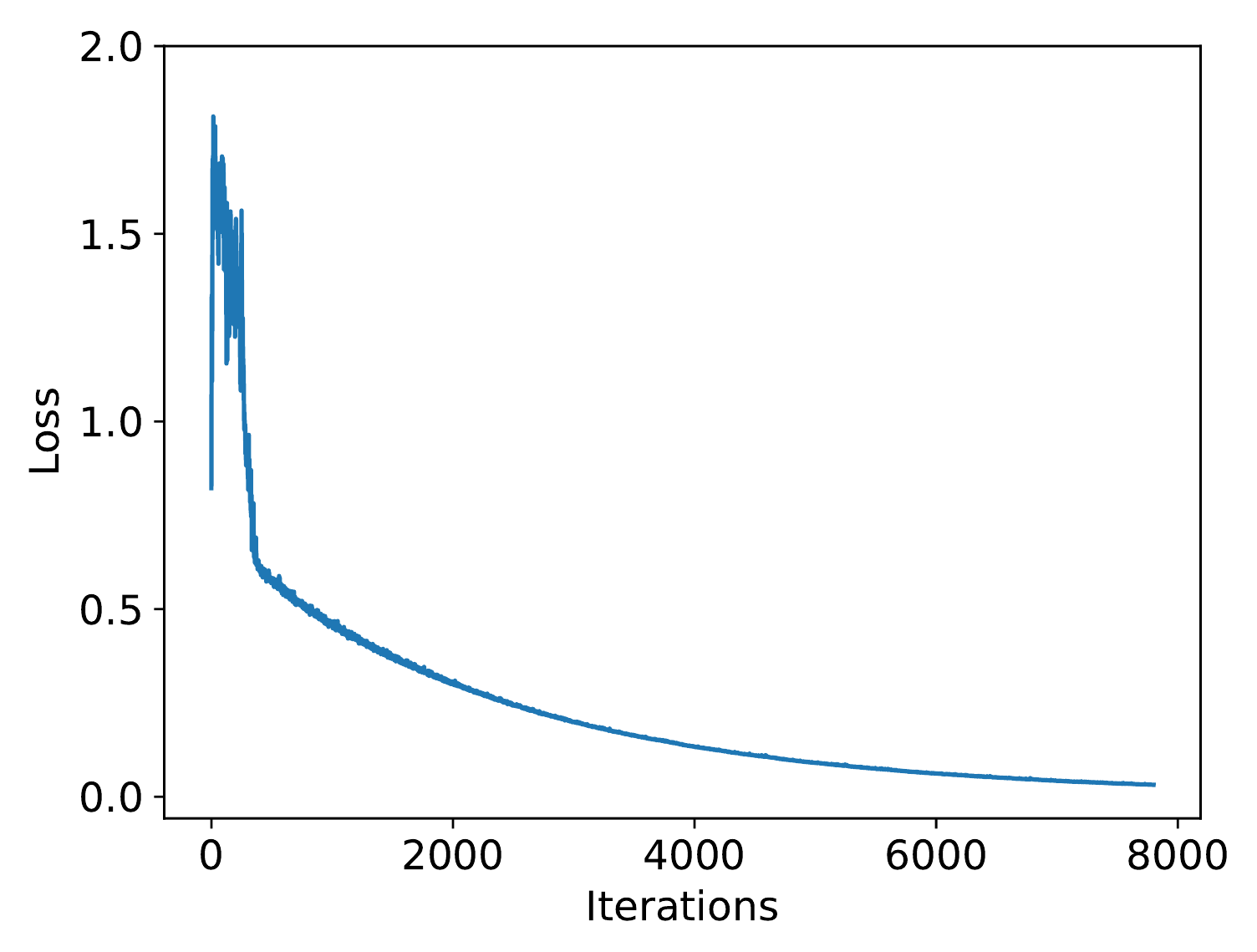} }}%
	\caption{Average training loss of five-fold cross-validation on the AFLW-112 and AFLW-64 dataset}%
	\label{fig:lossImages}%
\end{figure}

The proposed ResNet architectures were implemented using TensorFlow and trained on a Nvidia Tesla P100 GPU. The pre-processing, training and evaluation is implemented in one pipeline. The convolutional weights are initialized with the variance scaling initializer and trained with an initial learning rate of 0.1, which is decreased by the factor 10 after 30, 60, 80 and 90 epochs. The weight decay $\lambda$ of the L2 regularization excludes the loss of the batch normalization layers and has a value of 0.0002. We use a batch size of 256. All modified ResNets are trained separately for each angle.\\
\indent There are two training and testing procedures: (1) Five-fold cross-validation with the AFLW-112 and AFLW-64 dataset and (2) Five training-testing cycles with training on the whole AFLW-112 or AFLW-64 dataset and testing on the AFW-112 or AFW-64 dataset (see Fig.~\ref{fig:lossImages}). The ResNet18-112 and ResNet34-112 are trained for 200 epochs in both cases and the ResNet18-64 is trained for 120 epochs in case (1) and for 150 epochs in case (2). The epoch number was determined empirically.\\
\newpage
\indent The results are measured in mean absolute error (MAE) (see Eq.~\ref{eq:mae}), where $\hat{y}$ describes the predicted values and $y$ the true values in degrees. The number of testing examples is $n$.\\
\begin{equation}
\frac{1}{n}\sum_{i=1}^{n}|\hat{y}_i-y_i|
\label{eq:mae}
\end{equation}\\
\indent As in other approaches, the predicted and true values are mapped on discrete categories with a size of $15\degree$ (i.e. ...,]-7.5,7.5],]7.5,22.5],...) to predict the accuracy. If the predicted value is in the same category as the true value, the predicted value is classified as correct, otherwise as incorrect. A further applied evaluation method considers the mapped predicted and true values as true, if it matches the true category or the adjoining categories. This gives a range of $45\degree$, where the predicted value can be classified as correct.\\
\indent The mapping of the true and the predicted values on categories is problematic, because the cases where these values are located near the borders of the categories can distort the result. Furthermore, it is questionable, if the evaluation method with mapped categories ${\pm15\degree}$ error has high significance, as a wide range of degrees is considered as a correct prediction. The MAE on the other hand provides a clear interpretation of the results.
\subsection{Results}
\begin{figure}[h!]
	\centering
	\subfloat[ResNet18-112]{{\includegraphics[width=5.7cm]{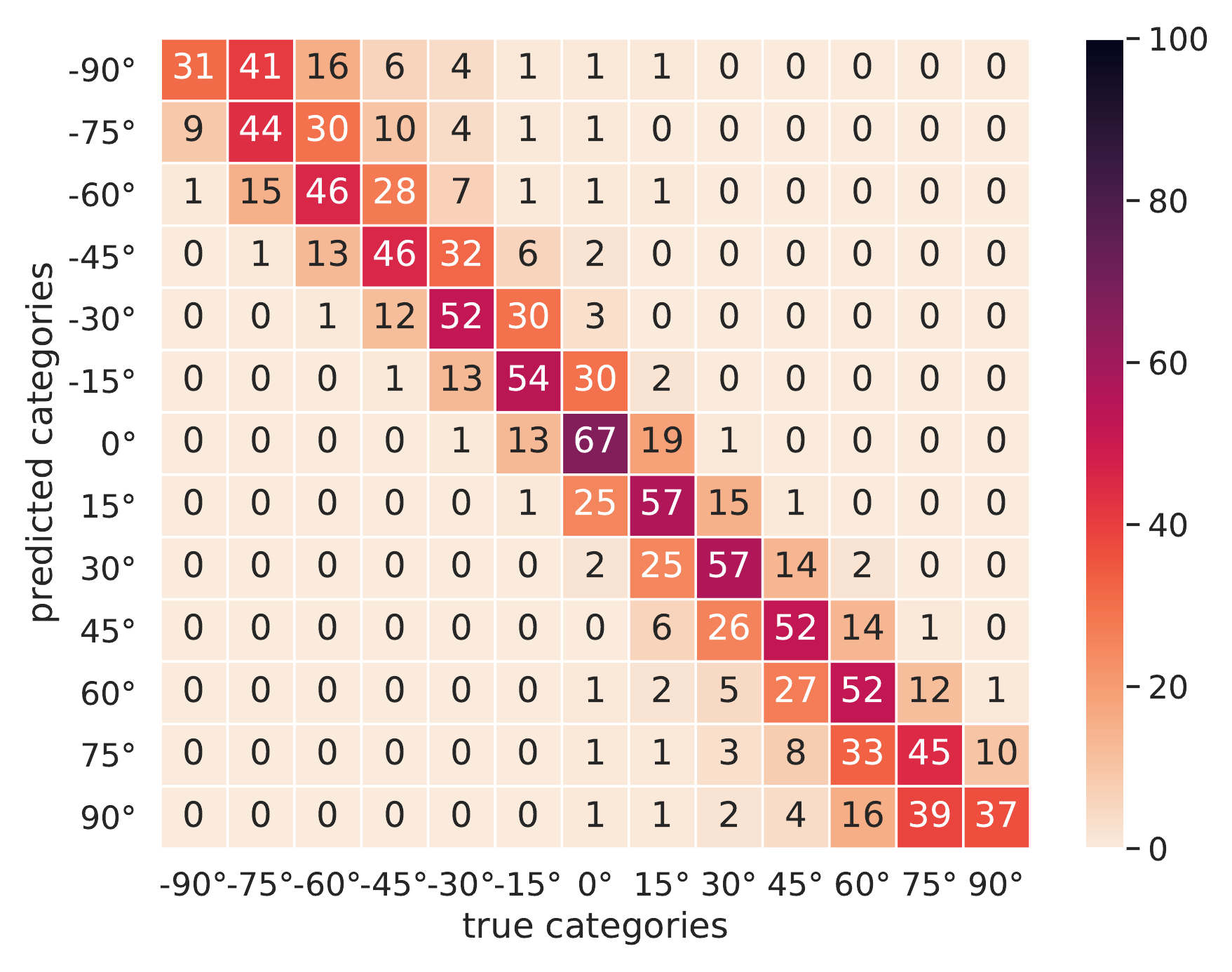}}}%
	\qquad
	\subfloat[ResNet18-64]{{\includegraphics[width=5.7cm]{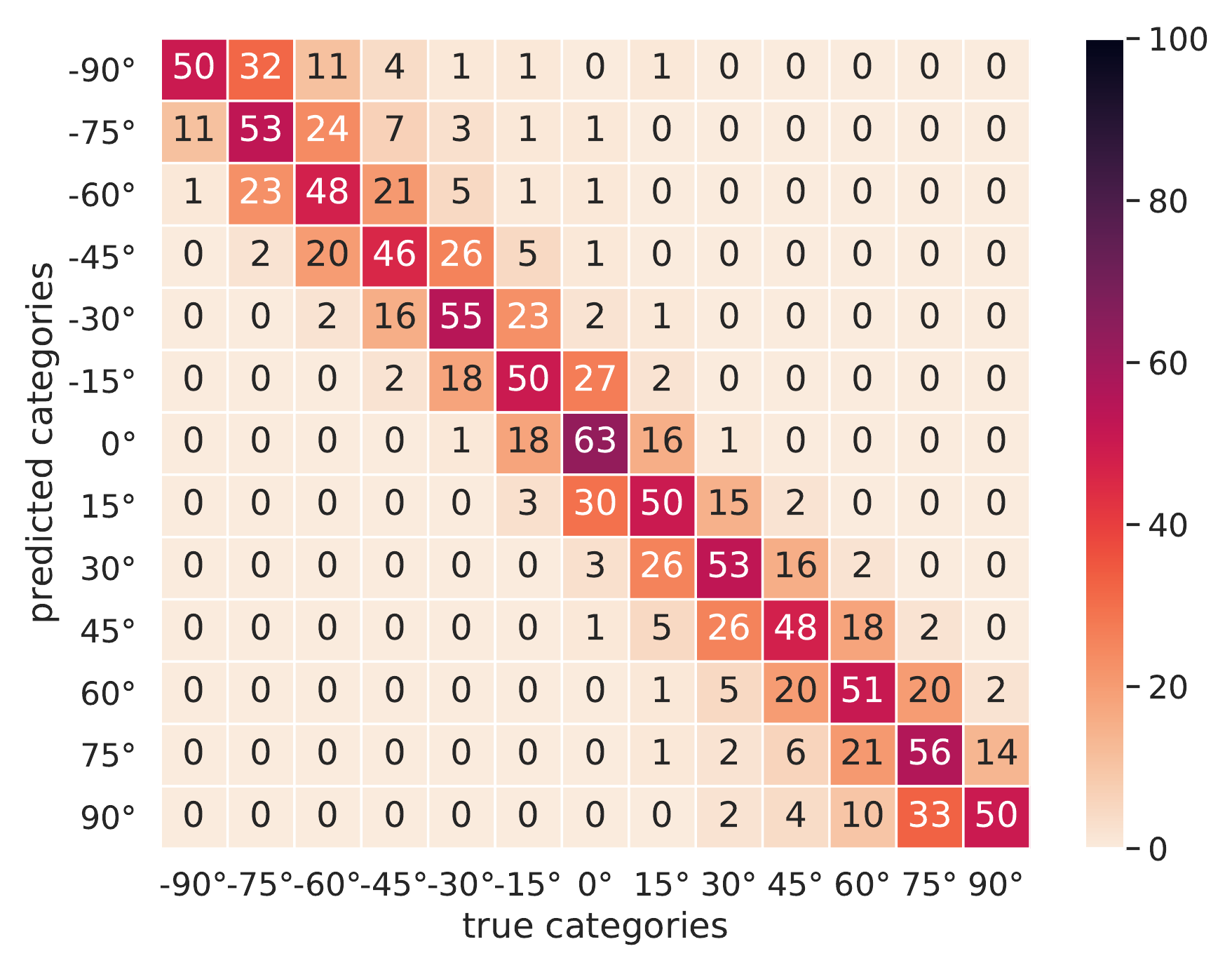} }}%
	\caption{These figures depict the average yaw angle heatmaps of a five-fold cross-validation on the AFLW-112 and AFLW-64 dataset. The distribution of images of each predicted category is in percentage.}%
	\label{fig:heatmaps}%
\end{figure}

Table \ref{Tab:resultResNetAFLW} shows the average training results for the three modified ResNets, evaluated on the pre-processed AFLW and AFW datasets with the methods explained in Section \ref{methods}. As presumed in the introduction, the comparison between the evaluated ResNets shows that their results for the three angles are quite similar. The results are similar, when tested on the AFLW dataset and when tested on the AFW dataset, only the ResNet34-112 shows worse results than the ResNet18-64 and ResNet18-112.\\
\indent Since the distribution of correctly classified images across label ranges is important for applications using head pose estimation, heatmaps are also considered as an evaluation tool. The heatmaps (see Fig.~\ref{fig:heatmaps}) show that the ResNet18-64 displays a more uniform distribution over the categories than the ResNet18-112. In comparison to the ResNet18-112, the ResNet18-64 shows a higher percentage of correctly classified images in categories closer to $\pm100\degree$ and a similar percentage of correctly classified images in categories closer to $0\degree$.
\begin{table}[h!]
	\caption{Average results of the modified ResNets: (1) tested with a five-fold cross-validation on the AFLW dataset and (2) tested on the AFW dataset in five training-testing cycles (marked with (AFW))}
	\begin{tabular}{|p{2 cm}|p{2.0cm}| p{2.0cm}|p{2.0cm}|p{3.5cm}|}
		\hline
		\textbf{Angle} & \textbf{MAE} & \textbf{Std. Dev.} &\textbf{Category} &\textbf{Category $\mathbf{\pm15\degree}$}\\
		\hline
		\hline
		\multicolumn{5}{|l|}{\small{\textbf{ResNet34-112}, CPU: 8 fps, GPU: 100 fps, 21.27 x $10^6$ parameters}}\\
		\hline
		\small{Yaw} & \small{$8.1\degree$} & \small{$\pm9.3\degree$} & \small{56.0\%} & \small{93.8\%}\\
		\small{Pitch} & \small{$6.2\degree$} & \small{$\pm5.4\degree$} & \small{61.8\%} & \small{97.8\%}\\
		\small{Roll} & \small{$3.8\degree$} & \small{$\pm3.8\degree$} & \small{78.5\%} & \small{99.8\%}\\
		\small{Yaw (AFW)} & \small{$15.6\degree$} & \small{$\pm15.8\degree$} & \small{38.0\%} & \small{77.9\%}\\
		\hline
		\hline
		\multicolumn{5}{|l|}{\small{\textbf{ResNet18-112}, CPU: 17 fps, GPU: 142 fps, 11.17 x $10^6$ parameters}}\\
		\hline
		\small{Yaw} & \small{$8.4\degree$} & \small{$\pm9.4\degree$} & \small{54.0\%} & \small{93.3\%}\\
		\small{Pitch} & \small{$6.0\degree$} & \small{$\pm5.3\degree$} & \small{62.4\%} & \small{98.1\%}\\
		\small{Roll} & \small{$3.8\degree$} & \small{$\pm3.7\degree$} & \small{78.4\%} & \small{99.8\%}\\
		\small{Yaw (AFW)} & \small{$13.5\degree$} & \small{$\pm15.3\degree$} & \small{42.6\%} & \small{83.8\%}\\
		\hline
		\hline
		\multicolumn{5}{|l|}{\textbf{ResNet18-64}, \textbf{CPU: 50 fps}, \textbf{GPU: 250 fps}, \textbf{4.25 x} $\mathbf{10^6}$ \textbf{parameters}}\\
		\hline
		\small{Yaw} & \small{$8.5\degree$} & \small{$\pm8.9\degree$} & \small{53.4\%} & \small{93.4\%}\\
		\small{Pitch} & \small{$6.5\degree$} & \small{$\pm5.5\degree$} & \small{59.6\%} & \small{97.7\%}\\
		\small{Roll} & \small{$3.9\degree$} & \small{$\pm3.8\degree$} & \small{77.8\%} & \small{99.7\%}\\
		\small{Yaw (AFW)} & \small{$13.2\degree$} & \small{$\pm13.3\degree$} & \small{41.1\%} & \small{83.3\%}\\
		\hline
	\end{tabular}
	\label{Tab:resultResNetAFLW}
\end{table}\\
\indent The parameter number decreases with the reduction of the model's complexity (see Table \ref{Tab:resultResNetAFLW}). As expected, the ResNet18-64 with 18 layers and an input image size of $64$ x $64$ pixels has the lowest number of parameters. The inference time was measured once on an Intel Core i7-6700 CPU running at 3.40GHz and once on this CPU equipped with a NVIDIA GeForce GTX 1060 6GB GPU. The frames per second (fps) rates of the different ResNets show a significant speed-up with the reduction of the model complexity and resulting decrease of parameters. The ResNet18-64 achieves 50 fps on the CPU, which is suitable for most real-time applications.
\begin{table}[ht]
	\caption{Results of the modified networks on the AFLW dataset in MAE and on the AFW dataset with $\pm15\degree$ error tolerance}
	\begin{tabular}{|p{3.5cm}| p{1.2cm}|p{1.2cm}|p{1.2cm}|p{1.2cm}|p{2.0cm}|p{1.2cm}|}
		\hline
		\multirow{2}{*}{\textbf{Approach}} & \multicolumn{3}{|c|}{\textbf{AFLW}} & {\textbf{AFW}} &\multirow{2}{*}{\textbf{Parameters}} & \multirow{2}{*}{\textbf{CPU}}\\
		\cline{2-5}
		& \textbf{Yaw} & \textbf{Pitch} & \textbf{Roll} & \textbf{Yaw} & &  \\
		\hline
		\small{\textbf{ResNet50 \cite{ruiz2018fine}}} &$\textbf{6.3\degree}$ &  $\textbf{5.9\degree}$ & $\textbf{3.8\degree}$ &  $\textbf{96.2\%}$ & {24.0 x ${10^6}$} & -\\
		\small{\textbf{LeNet-5 variant \cite{patacchiola2017head}}} & \small{$9.5\degree$} &  \small{$6.8\degree$} & \small{$4.2\degree$} & \small{$75.3\%$} &{4.6 x ${10^6}$} & 17 fps\\
		\small{\textbf{ResNet18-64 (ours)}} & \small{$8.5\degree$} &  \small{$6.5\degree$} & \small{$3.9\degree$} &  \small{$83.3\%$} &\small{\textbf{4.25} x $\mathbf{10^6}$} & \textbf{50 fps}\\
		\hline
	\end{tabular}
	\label{Tab:resultComparisonwithParams}
\end{table}\\
\newpage
\indent Other approaches evaluated on the AFLW and AFW datasets are summarized in Table \ref{Tab:resultComparisonwithParams}. The number of parameters of \cite{ruiz2018fine} is based on their provided open source implementation, which is executable on a GPU based system.\footnote{\url{https://github.com/natanielruiz/deep-head-pose}, accessed 09.01.2019.} In order to compare the frame rate, we reimplemented the LeNet-5 variant of \cite{patacchiola2017head}. In comparison, our ResNet18-64 has the lowest number of parameters while predicting more accurately than the LeNet-5 variant \cite{patacchiola2017head} and nearly as accurate as the ResNet50 \cite{ruiz2018fine}. Patacchiola and Cangelosi \cite{patacchiola2017head} also use low-resolution images with $64$ x $64$ pixels, while Ruiz et al. \cite{ruiz2018fine} take larger images with $224$ x $224$ pixels. To improve the computational efficiency, we believe that low-resolution images are better suited for real-world applications. Compared to our reimplementation of the LeNet-5 variant, our ResNet18-64 achieves a significantly higher frame rate on our CPU setup. Overall, our parameter-reduced ResNet18-64 achieves state-of-the-art precision and at the same time real-world applicability, even on CPUs.\\
\section{Conclusion and Future Work}
In this paper, we explored parameter-reduced Residual Networks (ResNets) of varying complexity for head pose estimation in order to achieve real-time performance. Based on the presumption that not all residual blocks contribute equally to the learning process, we showed that it is possible to reduce the number of parameters of the ResNet architecture while maintaining the performance. We proposed two new ResNet architectures for inputs of $112$ x $112$ pixels, one with 18 layers and one with 34 layers. To reduce the number of parameters even further, we proposed the ResNet18-64 with 18 layers for low resolution inputs of $64$ x $64$ pixels. The ResNet18-64 achieves real-time capability even on a CPU based system with a performance close to state-of-the-art results. To ensure real-world applicability, we evaluated the modified ResNets on the two in-the-wild datasets AFLW and AFW. In the future, it is possible to extend this approach to a model evaluating all three angles at once.


\begin{thebibliography}{8}
\bibitem{benenson2014ten}
Benenson, R., Omran, M., Hosang, J., Schiele, B.: Ten years of pedestrian detection,
what have we learned? In: Agapito, L., Bronstein, M.M., Rother, C. (eds.)
ECCV 2014. LNCS, vol. 8926, pp. 613–627. Springer, Cham (2015)

\bibitem{beymer1994face}
Beymer, D.: Face recognition under varying pose. In: CVPR. vol.~94, p.~137.
Citeseer (1994)

\bibitem{dementhon1995model}
Dementhon, D.F., Davis, L.S.: Model-based object pose in 25 lines of code. Int. J.
Comput. Vision \textbf{15}(1–2), 123–141 (1995)

\bibitem{diebel2006representing}
Diebel, J.: Representing attitude: Euler angles, unit quaternions, and rotation
vectors. Matrix  \textbf{58}(15-16),  1--35 (2006)

\bibitem{fanelli2013random}
Fanelli, G., Dantone, M., Gall, J., Fossati, A., Van~Gool, L.: Random forests
for real time 3{D} face analysis. International Journal of Computer Vision
\textbf{101}(3),  437--458 (2013)

\bibitem{fanelli2011real}
Fanelli, G., Gall, J., Van~Gool, L.: Real time head pose estimation with random
regression forests. In: CVPR 2011. pp. 617--624. IEEE (2011)

\bibitem{ferrario2002active}
Ferrario, V.F., Sforza, C., Serrao, G., Grassi, G., Mossi, E.: Active range of
motion of the head and cervical spine: a three-dimensional investigation in
healthy young adults. Journal of Orthopaedic Research  \textbf{20}(1),
122--129 (2002)

\bibitem{friesen1978facial}
Friesen, E., Ekman, P.: Facial action coding system: a technique for the
measurement of facial movement. Palo Alto  (1978)

\bibitem{geronimo2010survey}
Geronimo, D., Lopez, A.M., Sappa, A.D., Graf, T.: Survey of pedestrian
detection for advanced driver assistance systems. IEEE Transactions on
Pattern Analysis and Machine Intelligence  \textbf{32}(7),  1239--1258 (2010)

\bibitem{he2016deep}
He, K., Zhang, X., Ren, S., Sun, J.: Deep residual learning for image
recognition. In: Proceedings of the IEEE Conference on Computer Vision and
Pattern Recognition. pp. 770--778 (2016)

\bibitem{he2016identity}
He, K., Zhang, X., Ren, S., Sun, J.: Identity mappings in deep residual networks. In:
Leibe, B., Matas, J., Sebe, N., Welling, M. (eds.) ECCV 2016. LNCS, vol. 9908, pp.
630–645. Springer, Cham (2016)

\bibitem{hsu2018quatnet}
Hsu, H.W., Wu, T.Y., Wan, S., Wong, W.H., Lee, C.Y.: QuatNet: quaternionbased
head pose estimation with multi-regression loss. IEEE Trans. Multimedia
\textbf{21}(4), 1035–1046 (2018)

\bibitem{izard2013human}
Izard, C.E.: Human emotions. Springer Science \& Business Media (2013)

\bibitem{koestinger2011annotated}
Koestinger, M., Wohlhart, P., Roth, P.M., Bischof, H.: Annotated facial
landmarks in the wild: A large-scale, real-world database for facial landmark
localization. In: 2011 IEEE International Conference on Computer Vision
Workshops (ICCV Workshops). pp. 2144--2151. IEEE (2011)

\bibitem{kumar2017kepler}
Kumar, A., Alavi, A., Chellappa, R.: {KEPLER}: Keypoint and pose estimation of
unconstrained faces by learning efficient {H-CNN} regressors. In: 2017 12th
IEEE International Conference on Automatic Face and Gesture Recognition (FG
2017). pp. 258--265. IEEE (2017)

\bibitem{kuwahara2018driving}
Kuwahara, J., Nakazato, H.: Driving assistance system, US Patent 9,855,892, 2
January 2018

\bibitem{leach2014detecting}
Leach, M.J., Baxter, R., Robertson, N.M., Sparks, E.P.: Detecting social groups
in crowded surveillance videos using visual attention. In: Proceedings of the
IEEE Conference on Computer Vision and Pattern Recognition Workshops. pp.
461--467 (2014)

\bibitem{lecun2015deep}
LeCun, Y., Bengio, Y., Hinton, G.: Deep learning. Nature  \textbf{521}(7553),
~436 (2015)

\bibitem{lecun1998gradient}
LeCun, Y., Bottou, L., Bengio, Y., Haffner, P.: Gradient-based learning applied
to document recognition. Proceedings of the IEEE  \textbf{86}(11),
2278--2324 (1998)

\bibitem{lepetit2005monocular}
Lepetit, V., Fua, P., et~al.: Monocular model-based 3{D} tracking of rigid
objects: A survey. Foundations and Trends{\textregistered} in Computer
Graphics and Vision  \textbf{1}(1),  1--89 (2005)

\bibitem{leroy2013second}
Leroy, J., Rocca, F., Mancas, M., Gosselin, B.: Second screen interaction: an
approach to infer {TV} watcher's interest using 3{D} head pose estimation.
In: Proceedings of the 22nd International Conference on World Wide Web. pp.
465--468. ACM (2013)

\bibitem{li2014central}
Li, D., Pedrycz, W.: A central profile-based 3{D} face pose estimation. Pattern
Recognition  \textbf{47}(2),  525--534 (2014)

\bibitem{li2000support}
Li, Y., Gong, S., Liddell, H.: Support vector regression and classification
based multi-view face detection and recognition. In: Proceedings Fourth IEEE
International Conference on Automatic Face and Gesture Recognition (Cat. No.
PR00580), pp. 300--305. IEEE (2000)

\bibitem{niyogi1996example}
Niyogi, S., Freeman, W.T.: Example-based head tracking. In: Proceedings of the
Second International Conference on Automatic Face and Gesture Recognition.
pp. 374--378. IEEE (1996)

\bibitem{patacchiola2017head}
Patacchiola, M., Cangelosi, A.: Head pose estimation in the wild using
convolutional neural networks and adaptive gradient methods. Pattern
Recognition  \textbf{71},  132--143 (2017)

\bibitem{van2011head}
van~der Pol, D., Cuijpers, R.H., Juola, J.F.: Head pose estimation for a
domestic robot. In: Proceedings of the 6th Conference on Human-Robot
Interaction. pp. 277--278. ACM (2011)

\bibitem{ruiz2018fine}
Ruiz, N., Chong, E., Rehg, J.M.: Fine-grained head pose estimation without
keypoints. In: Proceedings of the IEEE Conference on Computer Vision and
Pattern Recognition Workshops. pp. 2074--2083 (2018)

\bibitem{schiele1995gaze}
Schiele, B., Waibel, A.: Gaze tracking based on face-color. In: International
Workshop on Automatic Face and Gesture Recognition. vol.~476. University of
Zurich Department of Computer Science Multimedia Laboratory (1995)

\bibitem{stiefelhagen2004estimating}
Stiefelhagen, R.: Estimating head pose with neural networks-results on the
{Pointing04 ICPR} workshop evaluation data. In: Proceedings of Pointing 2004
Workshop: Visual Observation of Deictic Gestures. vol.~1 (2004)

\bibitem{szegedy2015going}
Szegedy, C., et al.: Going deeper with convolutions. In: Proceedings of the IEEE
Conference on Computer Vision and Pattern Recognition, pp. 1–9 (2015)

\bibitem{veit2016residual}
Veit, A., Wilber, M.J., Belongie, S.: Residual networks behave like ensembles
of relatively shallow networks. In: Advances in Neural Information Processing
Systems. pp. 550--558 (2016)

\bibitem{wu2018simultaneous}
Wu, H., Zhang, K., Tian, G.: Simultaneous face detection and pose estimation
using convolutional neural network cascade. IEEE Access  \textbf{6},
49563--49575 (2018)

\bibitem{zhang2018cross}
Zhang, W., et al.: Cross-cascading regression for simultaneous head pose estimation
and facial landmark detection. In: Zhou, J., et al. (eds.) CCBR 2018. LNCS, vol.
10996, pp. 148–156. Springer, Cham (2018).

\bibitem{zhu2012face}
Zhu, X., Ramanan, D.: Face detection, pose estimation, and landmark
localization in the wild. In: 2012 IEEE Conference on Computer Vision and
Pattern Recognition. pp. 2879--2886. IEEE (2012)

\end{thebibliography}
\end{document}